\documentclass{IOS-Book-Article}

\usepackage{mathptmx}
\usepackage{soul}\setuldepth{article}
\usepackage{graphicx}
\usepackage[square,comma,sort&compress]{natbib}
\setcitestyle{numbers} 
\def\hb{\hbox to 11.5 cm{}}

\begin{document}

\pagestyle{headings}
\def\thepage{}

\begin{frontmatter}              % The preamble begins here.

%\pretitle{Pretitle}
\title{Real-Time Idling Vehicles Detection Using Combined Audio-Visual Deep Learning}

\markboth{}{July 2023\hb}
%\subtitle{Subtitle}

\author[A]{\fnms{Xiwen} \snm{Li}%
\thanks{Corresponding Author: Xiwen Li,  Scientific Computing and Imaging Institute, University of Utah, Sale Lake City, 84112, United States; Email: xiwen.li@utah.edu}},
\author[B]{\fnms{Tristalee} \snm{Mangin}},
\author[A]{\fnms{Surojit} \snm{Saha}},
\author[A]{\fnms{Rehman} \snm{Mohammed }},
\author[B]{\fnms{Evan} \snm{Blanchard}},
\author[B]{\fnms{Dillon} \snm{Tang}},
\author[B]{\fnms{Henry} \snm{Poppe}},
\author[B]{\fnms{Nathan} \snm{Searle}},
\author[C]{\fnms{Ouk} \snm{Choi}},
\author[B]{\fnms{Kerry} \snm{Kelly}},
\author[A]{\fnms{Ross} \snm{Whitaker}}%

\runningauthor{B.P. Manager et al.}
\address[A]{Scientific Computing and Imaging Institute, University of Utah, USA}
\address[B]{Department of Chemical Engineering, University of Utah, USA}
\address[C]{Department of Electronics Engineering, Incheon National University,
Republic of Korea}

\begin{abstract}
Combustion vehicle emissions contribute to poor
air quality and release greenhouse gases into the atmosphere,
and vehicle pollution has been associated with numerous
adverse health effects. Roadways with extensive waiting and/or
passenger drop-off, such as schools and hospital drop-off zones,
can result in a high incidence and density of idling vehicles. This
can produce micro-climates of increased vehicle pollution. Thus,
the detection of idling vehicles can be helpful in monitoring
and responding to unnecessary idling and be integrated into
real-time or off-line systems to address the resulting pollution.
In this paper, we present a real-time, dynamic vehicle idling
detection algorithm. The proposed idle detection algorithm and
notification rely on an algorithm to detect these idling vehicles.
The proposed method relies on a multisensor, audio-visual,
machine-learning workflow to detect idling vehicles visually
under three conditions: moving, static with the engine on, and
static with the engine off. The visual vehicle motion detector
is built in the first stage, and then a contrastive-learning-based
latent space is trained for classifying static vehicle engine sound.
We test our system in real-time at a hospital drop-off point in
Salt Lake City. This in situ dataset was collected and annotated,
and it includes vehicles of varying models and types. The
experiments show that the method can detect engine switching
on or off instantly and achieves 71.02 average precision
(AP) for idle detection and 91.06 for engine off detection.
\end{abstract}

\end{frontmatter}
\markboth{July 2023\hb}{July 2023\hb}

\section{Introduction}
 Poor air quality negatively impacts human health and was globally responsible for 6.5 million deaths and 21 billion in healthcare costs in 2015 \cite{/content/publication/9789264257474-en(A)}.  In particular, vehicle pollution has been associated with numerous adverse health effects, such as reduced cognitive function, cancer, and poor reproductive outcomes  \cite{Annavarapu2016CognitiveDI(B), Rice2016LifetimeET(C), Lewtas2007AirPC(U)}. Likewise, idling vehicles are significant contributors to greenhouse gas emissions \cite{EPA_2023(G)}.
 Roadways where idling vehicles tend to congregate, such as schools and hospital drop-off zones, can produce micro-climates of increased vehicle pollution \cite{Ryan2013TheIO(H)}.  Populations that are particularly vulnerable to vehicle pollution include children and individuals in wheelchairs since their breathing height is closer to the height of combustion exhaust \cite{Sharma2018ARO(D),Kenagy2015GreaterND(E)}. Idling vehicles 
are especially problematic in confined locations such as underground mines \cite{Mischler2010ControllingAM}. Moreover, idling among fleets, such as long-haul idling trucks at depot/delivery centers, causes excess operational costs due to wasted fuel and engine wear \cite{GEOTAB, CanadianMining}.  

 The detection or monitoring of idling vehicles can impact policies that can subsequently reduce pollution.  For instance, research has found that while conventional, static, anti-idling signage and education campaigns have mixed results, anywhere from little effect on driver behavior to improving air quality \cite{Mendoza2022AirQA, Ryan2013TheIO(H), Eghbalnia2013ACP(I), Meleady2017SurveillanceOS(J), Vyver2018MotivatingTS(K),Dowds2013ComparisonsOD(L),Carrico2009CostlyMA(M)}, dynamic signage may have a greater impact.  For instance, behavioral research has found that dynamic radar-based speed displays are more effective than static signage at reducing vehicle speeds \cite{Winnett2003VEHICLEACTIVATEDS(N), doi:10.1177/0361198105191800112(O),Lee2006EffectivenessOS(P),Cruzado2009EvaluatingEO(Q),Stern2010EvaluationDD(R),Ando2017LongtermEA(S),Gehlert2012EvaluationOD(T)}. 

This paper focuses on the design and evaluation of an idling vehicle detection (IVD) system that can be deployed in parking or drop-off areas in order to monitor and respond to driver behaviors.  
To the best of our knowledge, only one previous work \cite{Bastan2018IdlingCD} has discussed a method for IVD. That method relies on infrared imaging to detect heat from the engine block of a vehicle, which has several limitations, as described in the next section.  

\section{Related Work}

\subsection{Idling Vehicle Detection}
Previous work \cite{Bastan2018IdlingCD} has proposed to automatically detect IVs via the use of an infrared camera, which monitors the target area. They perform object detection on a heatmap. This method has several disadvantages: (1) The first is high latency because it takes time for heatmap to accumulate and dissipate. Also, many infrared cameras often have relatively low frame rates (e.g., one image every 5 seconds).  (2) According to our preliminary experiments, environments with direct sunlight or high ambient temperatures can adversely affect the model's ability to detect a hot engine block.  (3) Engine blocks and exhaust pipes (as noted by the authors \cite{Bastan2018IdlingCD}) are the main high heat area; however, false positives/negatives occur when the engine block faces away from the camera because the engine block is the main heat source. In addition, our preliminary results on detecting heat from vehicle exhaust (rear of car) are inclusive.  (4) Infrared imaging cameras are generally more expensive and not easily deployable in a wide variety of settings. The proposed method uses a common RGB webcam and wireless microphone array for video and audio data acquisition. Additionally, the problem is defined as an audio-visual workflow, including vehicle motion detection and audio classification.  The method uses real-time video and audio clips and can detect engines switching on or off in less than 1 second. Thus, the proposed audio-visual approach represents an attractive alternative to infrared-based IVD.

\subsection{Video Understanding}
Video understanding is an important area of study within the field of computer vision. Action detection, often the main task of video understanding, typically estimates a 2D bounding box (region in the image) and label for each action event on video frames. Convolution neural networks (CNNs) have been successfully applied to vehicle motion detection, as in \cite{STSN_ECCV2018, Feichtenhofer2018SlowFastNF, HLSTM_CVPR2020, WOO_ICCV2021, 3DFCNN_2022}. With respect to autonomous driving, works using ego-vehicle cameras such as \cite{MODNet,Siam2018MODNetMA,Song2022MotionEF,Zhao2019RealTimeVM} focus on detecting vehicle motion. However, the problem statement in this paper requires surveillance-style camera placement, as in \cite{Bastan2018IdlingCD,Lopez2019ParkingLO}, for the purpose of monitoring the target area. The proposed design is vision guided, and we use the SoTA action detection model YOWO \cite{Kpkl2019YouOW} as the first stage for locating vehicles in the video frames.

\subsection{Audio Classification}
A great deal of research has addressed problems related to sound classification such as \cite{Kong2019PANNsLP, Saeed2020ContrastiveLO, Nasiri2021SoundCLRCL} and speech diarization \cite{speaker_diarization_2007}. Devices such as microphone arrays and acoustic cameras \cite{Zunino2015SeeingTS} can help localize sound sources relative to a visual frame of reference. We have evaluated microphone arrays and beamforming algorithms to separate vehicle engine sound in each direction. Nevertheless, from empirical evaluation, we have observed that the beamforming technique cannot reliably resolve a single vehicle in outdoor setup because of the far-field attenuation and the ambient noise, and the localization accuracy is poor.
%However, we have found, through experimentatino,  that the beamforming approach could not reliably resolve a single vehicle outdoors, because the ambient noise, required base-lines, and synchronization made localization at the reqiured distances imprecise. 
Therefore, in order to receive a stable and clear audio signal of vehicle engines, we place individual wireless microphones in a row at the roadside, as shown in Fig. \ref{fig:sys_setup}, and use machine learning models to identify idling engines. Due to access to limited training data, we resort to models built using unsupervised learning techniques, such as contrastive learning \cite{simCLR_2020, MoCo_2020}, on publicly available audio datasets \cite{UrbanSound8K_2014, ESC-50_2015, AUDIOSET_2017}. %and use machine learning adapted to limited training data to identify engine sounds and their proximity to microphones.

\subsection{Audio-Visual Learning}
Audio-visual learning is an emerging topic in computer vision. Several works, such as \cite{Afouras2021SelfsupervisedOD, Owens2018AudioVisualSA, Audio_visual_2021, Audio_visual_2018}, learn audio-visual, co-occurrence features. However, our IVD problem cannot be resolved by feature co-occurrence because a stationary vehicle has visual presence but lacks audio presence. Furthermore, our experiments show that vehicles are accurately identified in videos, regardless of their engine status.  Thus, we have opted for a vision-guided system design, rather than pure, mixed audio-visual recognition. 

\begin{figure}
\centering
\begin{tabular}{c}
\includegraphics[width=\linewidth]{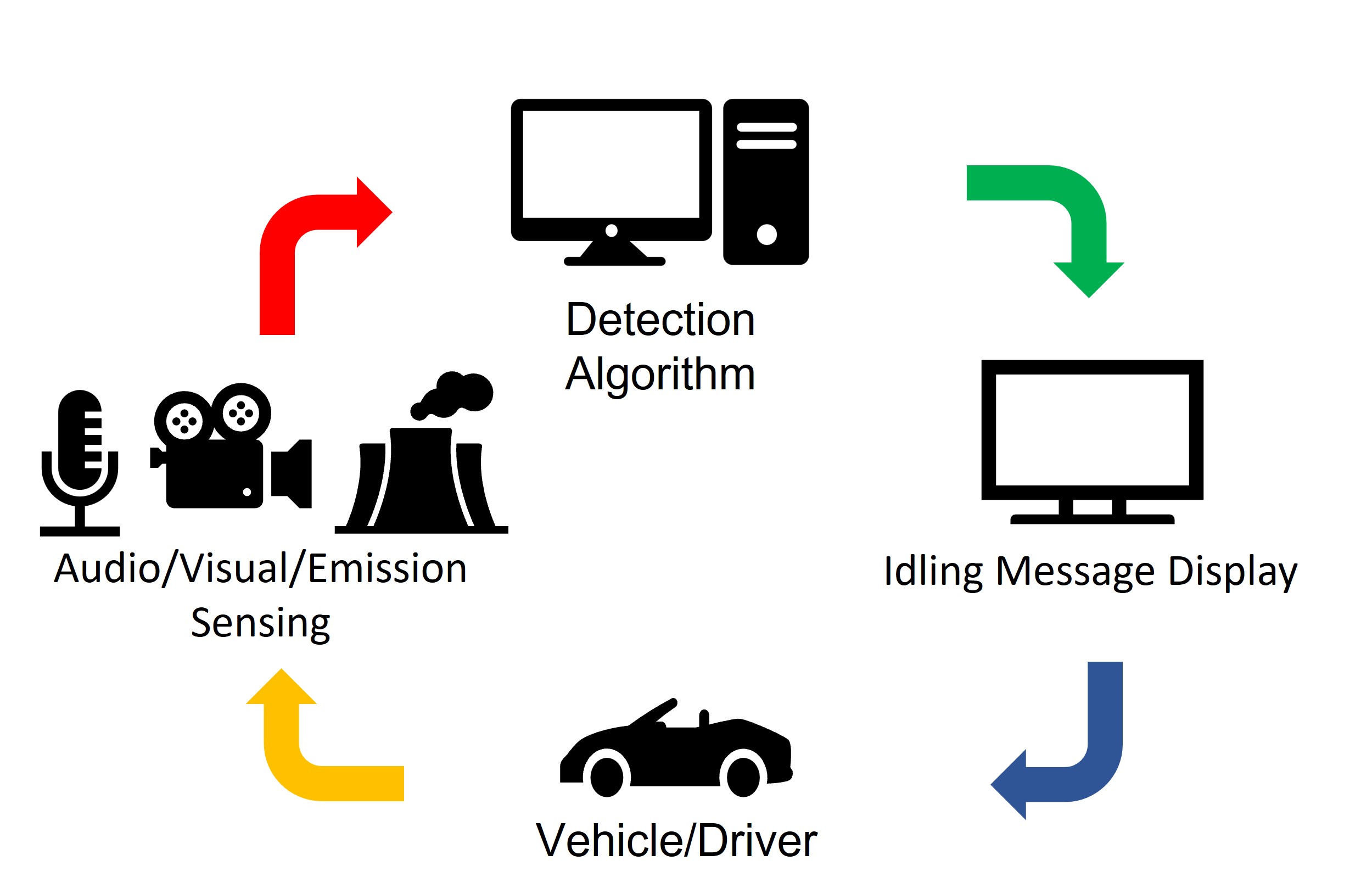} \\ 
% (a) \\
% \includegraphics[width=\linewidth, height=150px]{imgs/message.png} \\
% (b) \\
\end{tabular}
\caption{Proposed System Design. The yellow arrow collects vehicle motion, engine sound, and pollutant concentrations. The red arrow represents data transmission to the computer. The green arrow denotes sending the predicted idling status to the displays. The blue arrow represents the driver receiving the information from the display and potentially making behavior changes. 
% (b) Dynamic Message Flowchart. Messages are displayed to drivers on two large outdoor screens according to this flowchart.  The message only shows whether or not a vehicle is idling in the area, even though the algorithm can detect the number of idling vehicles and, more specifically, which vehicles are idling.
}
\label{fig:SmartAir}
\end{figure}

\begin{figure*}[t]
  \centering
  \begin{tabular}{c}
  \includegraphics[width=1.2\linewidth] {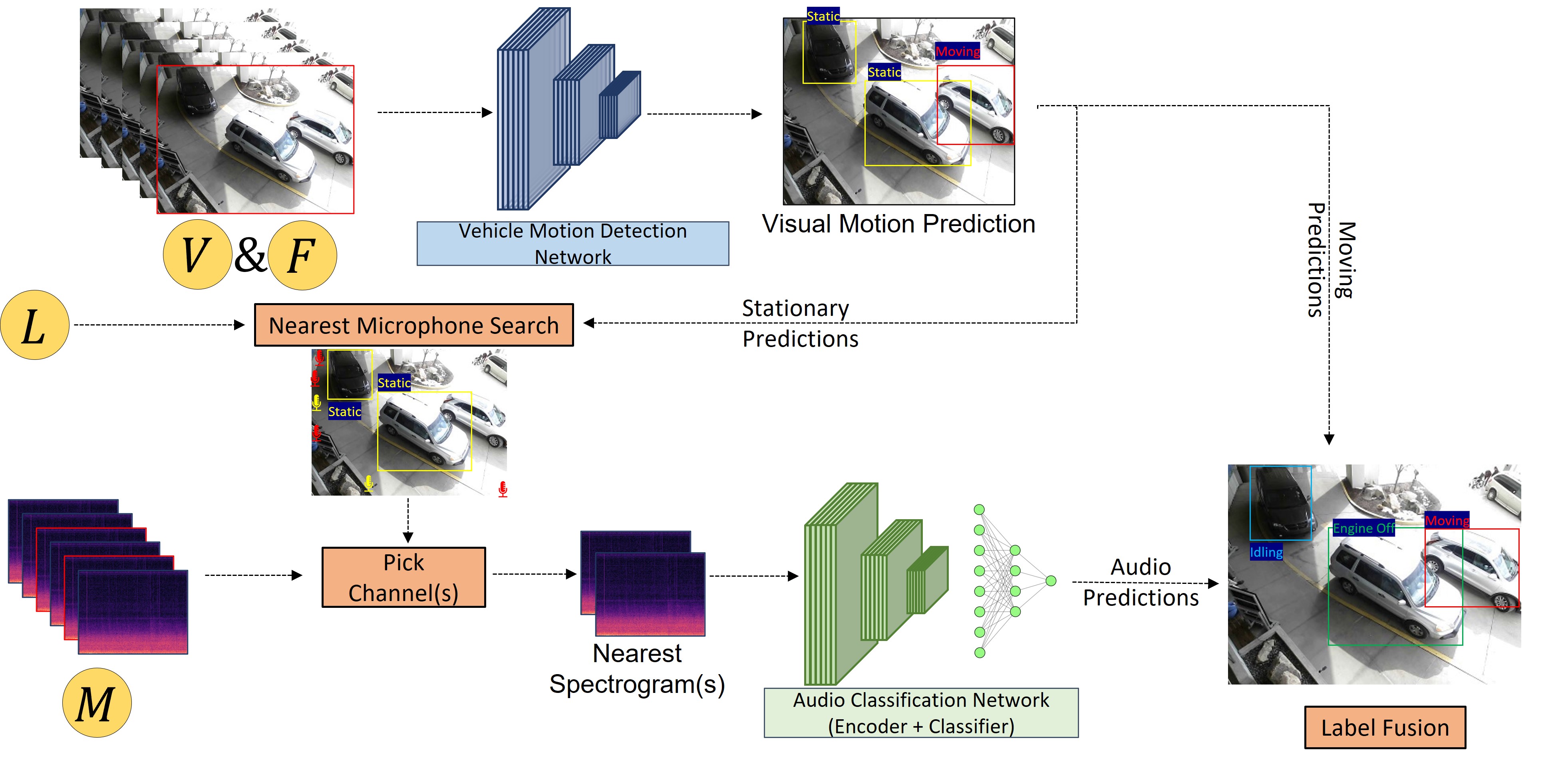} \\
  (a)\\
  \includegraphics[width=0.75\linewidth]{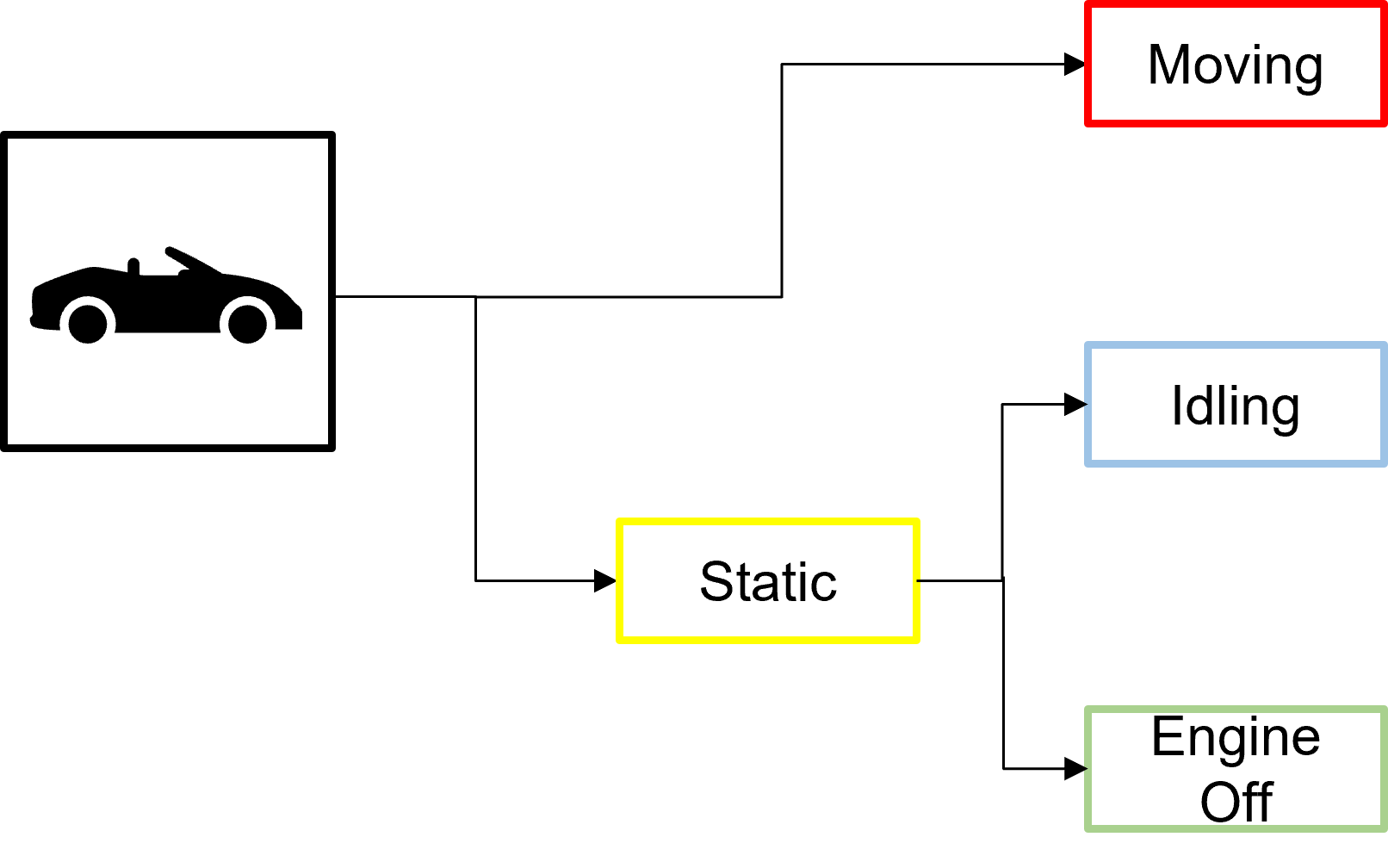}\\
  (b)
  \end{tabular}
  \caption{(a) Our IVD Algorithm. (b) Class Definition Hierarchy.}
  \label{fig:workflow}
\end{figure*}

\section{Method}
\subsection{Problem Definition}
We define the problem as localizing and deciding if multiple vehicles in a video clip are idling individually in a drop-off area. We define three vehicle status classes $Y\in \{Y_{\rm moving}, Y_{\rm off}, Y_{\rm idling}\}$ and show them in Fig. \ref{fig:workflow}(b):
\begin{itemize}
\item \textbf{Moving}. A vehicle is moving.
\item \textbf{Engine Off}. A vehicle is stationary with the engine off.
\item \textbf{Idling}. A vehicle is stationary with the engine on.
\end{itemize}
By definition, an electric vehicle is either moving or off.  
We define the problem in an audio-visual manner. To be specific, given a video clip $V\in R^{D\times H\times W\times C}$ visually containing vehicles $v_{1}, v_{2},...,v_{n}$, microphone pixel location $L$ and audio clips $M\in R^{N_{c}\times SR}$ at the same moment, our model estimates a bounding box $BB^{v_{i}}$ and $Y^{v_{i}}$. $D$ is the number of video frames. $H$, $W$, and $C$ are the height, width, and channel number of an RGB frame. $N_{c}$ is the number of wireless microphones. $SR$ represents the microphone sample rate. The observed time-domain audio signal $M$ is defined as the addition between engine sound $S$ and environment noise $N$:
\begin{equation}
\label{audio_def}
    M=S+N (M,S,N\in R^{N_{c} \times SR})
\end{equation}
The model estimates class label $Y^{v_{i}}$ of vehicle $v_{i}$ given information of bounding box $BB$, motion label $Y_{motion}$ and nearest audio signal $M^{v_{i}}$:
\begin{equation}
P(Y^{v_{i}}|BB^{v_{i}}, Y_{motion}^{v_{i}}, M^{v_{i}})
\end{equation}

To solve this problem, we propose a two-stage visual-guided audio classification algorithm shown in Fig. \ref{fig:workflow}(a). A vehicle  detector is trained in the first stage using the SoTA video understanding model, which learns to detect a bounding box $BB^{v_{i}}$ and a moving or stationary label $Y_{motion}^{v_{i}}$ for each detected vehicle. For each stationary vehicle detected in the frame, we determined (through proximity in the image space) the closest microphone. In the second stage, the model classifies audio acquired from each of the closest microphones into sound with/without engine presence.

\subsection{System Setup}
\begin{figure}
    \centering
    \includegraphics[width=0.5\linewidth, height=200px]{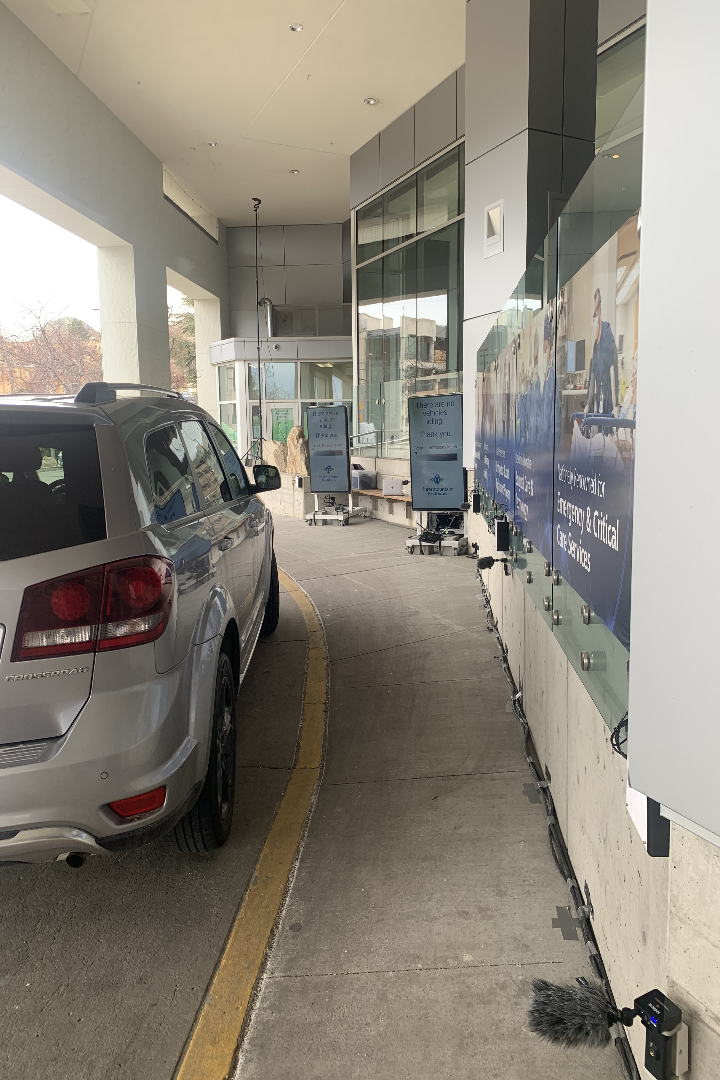}
    \caption{System Setup. Wireless microphones are stuck on the right wall. The camera is mounted on a tripod near rocks.}
    \label{fig:sys_setup}
\end{figure}

Our system setup at the test location is shown in Fig. \ref{fig:sys_setup}. We set up an RGB camera on a tripod at an elevation of approximately 20 feet, pointed toward the target area. For instance, at the hospital collection area, the RGB camera monitors a two-lane drop-off area, and the wireless microphone transmitters are evenly spaced along the roadside (at 2.6 meters intervals). The microphone transmitters send the acquired signals to the microphone receivers.  The microphone receivers are connected to the desktop computer, which collects the data and runs the algorithm. An EMMET C960 webcam was used for the video, and 3 Rode Wireless GO II sets with unidirectional microphones are used for the audio. Each Rode Wireless GO II set has one receiver and two transmitters/microphones, resulting in a total of six microphones.

\subsection{Vehicle Motion Detection}
The vehicle motion detector (as in Fig. \ref{fig:workflow}) estimates, for every detected vehicle, whether it is in motion $Y_{moving}$ or stationary $Y_{stationary}$. For achieving this, we use a deep-learning, video-based, object-detection model, YOWO\cite{Kpkl2019YouOW}, considered a state-of-the-art benchmark, which is a video-understanding extension of the (static) object detection model YOLO\cite{Redmon2016YOLO9000BF}. The neural network extracts 3D clip features using ResNeXt\cite{Xie2016AggregatedRT}, extracts 2D frame features using DarkNet\cite{Redmon2016YOLO9000BF}, and fuses 3D features and 2D features for action detection. YOWO's input is a video clip $V\in R^{D\times H\times W\times C}$ and the video clip’s last frame $F\in R^{H\times W\times C}$ of $V$. In general, YOWO learns and detects vehicles' motion on $F$ using $V$ and $F$.

\subsection{Nearest Microphone Search}
Once each vehicle's location and moving/stationary status is found, the algorithm finds the closest microphone to that vehicle. For this, the algorithm relies on image (pixel) locations for each microphone in the video frame, determined interactively by users at the time of system setup. In the setup, we develop a script that accepts a mouse click for each of the six microphone locations in the video frame and produces the pixel-microphone dictionary $L$ for that deployment.  The process takes approximately 1 minute.  Having the user input these locations at the beginning of each deployment ensured that the algorithm had the correct microphone pixel location, because the video camera angle can vary due to equipment takedown and installation each deployment. Each predicted $BB_{v_{i}}$ 's centroid is computed for the search. $L$ is stored and can be called during the search for the nearest microphone. The next step uses the audio channel from the nearest microphone for sound classification.

\subsection{Engine Sound Classification}
In the second stage, the algorithm classifies observed sound $M$ and environment noise $N$ in Eq. \ref{audio_def}. The model assumes that audio $M=N$ if no idling car shows up near the microphone, and $M=S+N$ otherwise. Usually, audio where an idling engine is present has power at a certain combination of frequencies relative to audio that consists entirely of background noise (e.g., wind, people talking, distant traffic). We refer the former as foreground signal and the later as background signal. 

There are two difficulties when building such a binary classifier in real time.  The first difficulty is that there is limited training data because labeling idling/non-idling ground truth is difficult and time-consuming, and would require too many hours of training data, making efficient deployment in new locations infeasible. However, the system requirements indicate that the classifier must generalize to new, previously unheard cars even with limited training data.  

The next difficulty is that the data has multiple outliers because the real-world deployment environment must deal with practical limitations on the placement of microphones, microphone cutoff, and environmental interference.  Some microphones are not placed in the optimum location for engine noise due to the presence of sidewalks, wheelchair access, or existing infrastructure. The placement of the microphones also impacts the signal quality that the transmitter has with the receiver. Additionally, sound events such as helicopters flying overhead or people talking right by the microphone can confuse the classifier.  Because of these issues, we have found that a simple frequency-selection-based power threshold is not reliable, and we have developed a machine-learning-based method applied to time-frequency data to differentiate $S$ and $N$.  Moreover, in practice, it is time-consuming to annotate (video) frame-by-frame idle ground truth, because an annotator must watch the video and refer to the audio back and forth to achieve frame-level accuracy. 

The problem of building classifiers with limited training data is an important and active area of machine-learning research, with some promising preliminary results. One effective strategy is to develop latent spaces using large amounts of unlabeled or weakly labeled training data, and then to leverage this latent space to the target problem with limited data.  In this light, we have developed a supervised contrastive learning approach for subsequent audio classification.  Supervised contrastive learning (SCL) (first proposed in the computer vision literature \cite{Khosla2020SupervisedCL} ) delineates interclass audio samples while alienating intraclass samples. Following architecture and SCL loss functions from \cite{Khosla2020SupervisedCL}, the architecture for the proposed system relies on a ResNet50 feature encoder, a fully connected projector, and a linear/nonlinear classifier. First, the encoder and projector are pretrained using SCL loss on a large public audio dataset ESC-50. Next, the pretrained encoder is frozen and fed a small amount of the labeled foreground/background data from the new site/deployment. We found that the on-site data in the SCL latent spaces is distributed in an easily separable fashion. Thus, we can use simpler classifier (with fewer degrees of freedom) to differentiate between foreground and background data.

For processing by the neural network, the time-domain, audio signal is converted to a 2D spectrogram through a short-time Fourier transform (STFT). Compared to a regular Fourier transform, STFT reflects the local frequency domain over time, computes magnitude, and concatenates them vertically into a spectrogram with shape $T\times F$, where $T$ is the number time steps, and $F$ is the number of frequency bins.  The spectrograms are encoded using a ResNet50, which is pretrained on ImageNet \cite{Khosla2020SupervisedCL}, and then fine tuned on ESC-50 \cite{Piczak2015ESCDF} using the SCL approach described above. Similar to \cite{Nasiri2021SoundCLRCL} and \cite{Saeed2020ContrastiveLO}, the model maps the input spectrogram to a normalized 2048 dimension vector and further projects it to a normalized 64 dimension latent vector on a hypersphere. A classifier is trained to classify 2048 dimension latent vectors.

In the last step, we keep $Y$ and $BB$ for moving vehicles. We replace $Y$ as either $Y_{idling}$ or $Y_{eoff}$ for stationary vehicles according to audio classifier's output.

\section{Experiments}
The system was deployed at the main entrance of the hospital test site for 14 test days. On average, about 10 vehicles are picking up and dropping off patients every hour, with some typically busier times (e.g., morning and mid afternoon). The vehicle motion detector, engine sound classifier, and entire IVD pipeline performance are evaluated qualitatively and quantitatively on a single day of held-out data.

\subsection{Dataset}
\subsubsection{Audio-Visual IVD Dataset}
We performed field tests at the test site for 14 days. To the best of our knowledge, there is no existing IVD dataset matching our setup. Because video and audio sample amounts differ significantly (foreground audio samples are very limited since drivers' idling time is much shorter than entire recording), they are collected separately on different days. Sampled every 1 second from the first 3-days 10-hour recordings, our video training set consists of 33015 clips, and our video validation set consists of 8252 clips. Our video test set has 13271 clips sampled from a random one test day recording from the remaining 11 days. All video data is annotated with bounding boxes and motion status. The camera fps is 25. The audio training set has 8721 foreground (positive) samples and 30618 background (negative) samples collected every 1 second from a random five days' first 80\% of recording. To evaluate the audio model's generalization ability over new vehicles, the audio validation set has 2491 positive samples and 8245 negative samples sampled from those 5 days' last 20\% of recording. The sample rate of the microphone is 48000Hz. All audio samples are 5 seconds long and centered at the video clip's last frame. They are input pairs for the algorithm. The data acquisition script synchronizes the video and audio. Notes are taken on-site for annotating the ground truth per frame.

\subsubsection{Urban Sound Classification} 
Our audio model is pretrained on ESC-50. ESC-50 \cite{Piczak2015ESCDF} is a public environment sound event classification dataset that contains 2000 audio recordings evenly distributed over 50 classes, including engine idling and common background noise sources similar to that observed at the test-site field deployments, such as wind, speech, and helicopters. In the ESC-50 dataset, each audio sample is 5 seconds long, and the samples are recordings from the public project Freesound.

\subsection{Experimental Setup}
The vehicle motion model is trained using an Adam optimizer with a learning rate of 0.0001 with batch size 20 on a Nvidia RTX4090 24GB GPU. Following \cite{Kpkl2019YouOW}, five anchors are precomputed on the training dataset. The audio model is trained using the same GPU and learning rate 0.0001 with batch size 128. The audio latent space dimension is 2048. Each input audio sample is 5 seconds long and is normalized. Since audio positive and negative samples are imbalanced, we manually balance them in each mini-batch for stochastic gradient descent.

\subsection{Metrics}
We evaluate our algorithm by computer vision's common object detection metrics: average precision (AP), mean average precision (mAP), and accuracy. These metrics also compute intersection over union (IoU), precision, and recall curve.
\subsubsection{Intersection Over Union (IoU)}
IoU measures how predicted bounding box $BB_{p}$ overlaps with ground truth box $BB_{gt}$. It is a common metric used for object detection in computer vision.
$$IoU=\frac{BB_{p}\displaystyle\cap BB_{gt}}{BB_{p}\displaystyle\cup BB_{gt}}$$

\subsubsection{Metrics for Quantitative Analysis}
In object detection, a true positive is defined as a detection with bounding-box IoU (relative to ground-truth bounding boxes) greater than a threshold, whereas a false positive is defined as a detection/output from the neural network with IoU smaller than a given threshold. A false negative is defined as a ground truth without a corresponding, overlapping, bounding box from the NN. For object detection tasks, typical choices for these thresholds are 0.5 and 0.75. With precision and recall computed from these, a Precision-Recall (PR) curve can be plotted to evaluate the performance of an object detector on a particular class. In audio classification, precision and recall are defined in the conventional manner for binary classifiers.   AP is the interpolated area under PR curve for each class. mAP is the average AP value over classes.

\subsection{Vehicle Motion Detection Performance}
We evaluate the performance of vehicle motion detection using mAP and AP. For test-site deployment, the evaluation focuses on the motion detector's generalization ability across test data. Because the tripod is redeployed each day, the camera pose varies slightly from day to day. The lighting conditions, traffic conditions, and types of vehicles also differ. For example, the variability of light conditions between sunny and cloudy days is significant. The motion detector was trained and validated on the first 3 train/validation days. Sampled every second, the first 3-day data is divided into 80\% training clips and 20\% validation clips, resulting in 33015 training clips and 8252 validation clips. Additionally, 13271 test clips are sampled in the same way for another random single day. Table \ref{tab:VDM} summarizes the performance comparison between validation set and test set. APs of stationary and moving are comparable for validation and test clips. For the same IoU threshold, the test data's stationary AP is about $15\%$ lower than validation data. Moving AP is even 9\% higher because, we believe, validation samples have fewer vehicles than test samples. However, from these results we can conclude that the trained vehicle motion detector is capable of localizing vehicle motion on a different single day's test data with 3 days of training data, even with different pose and lighting conditions.

\begin{table}[ht]
\caption{Vehicle Motion Detection Performance}
\label{tab:VDM}
\begin{center}
\begin{scriptsize}
\begin{tabular}{|c|c|c|c|c|}
\hline
 & data type & mAP (\%) & AP Stationary (\%) & AP Moving (\%) \\
\hline
IoU\textbf{@0.5} & validation & 84.49 & 95.83 & 73.15 \\
\hline
IoU\textbf{@0.75} & validation & 60.94 & 80.85 & 41.03 \\
\hline
IoU\textbf{@0.5} & test & 87.04 & 91.14 & 82.94\\
\hline
IoU\textbf{@0.75} & test & 48.22 & 64.50 & 31.94\\
\hline

\end{tabular}
\end{scriptsize}
\end{center}
\end{table}

\subsection{Engine Sound Classifier Performance}
\begin{figure}
    \centering
    \begin{tabular}{c|c}
    \includegraphics[width=0.45\linewidth]{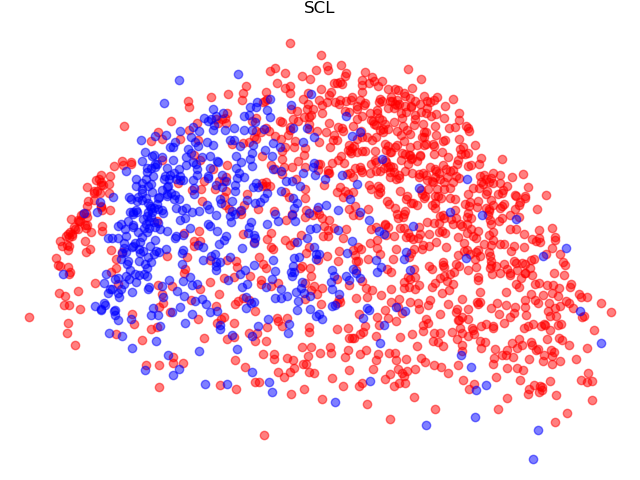} &
    \includegraphics[width=0.45\linewidth]{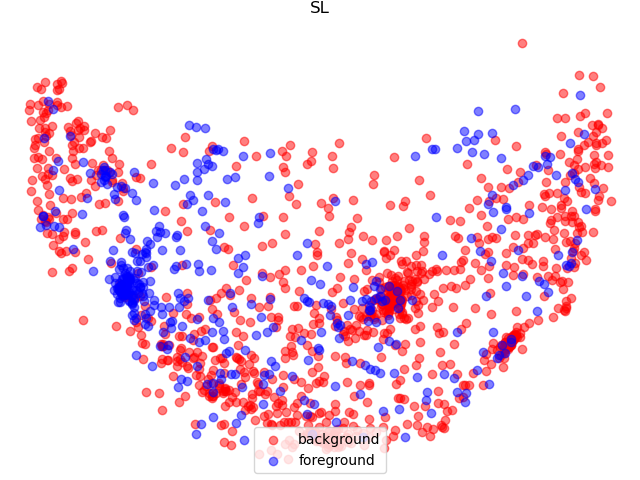}
    \end{tabular}
    \caption{MDS 2D Visualization on Audio Encoder Latent Space. Due to the huge amount of validation samples, we feed part of our validation data into trained SCL and SL's encoder. The left side shows projected latent space of SCL in 2D dimension. The right side shows SL's latent space. Red dots are background samples. Blue dots are foreground samples.}
    \label{fig:scl_vs_sl}
\end{figure}

The sound classifier is evaluated using precision, recall, and F-score. Groundtruth static vehicles are selected, and the respective ground truth bounding boxes are fixed to find the closest microphone to cut audio samples from for training and testing. The effectiveness of supervised contrastive learning (SCL) and supervised learning (SL) is compared by visualizing validation data latent vectors in normalized 2048 dimension space, as seen in Fig. \ref{fig:scl_vs_sl}. This visualization shows that in the SCL latent space positive and negative vectors are clustered and potentially separable. To simulate nonlinearity, we chose a two-layer perceptron with 1024 internal nodes with a relu activation as the classifier on the latent space. After training, the classifier's performance on validation data was computed and is shown in Table \ref{tab:audio_acc}. Although both have similar recalls, SCL's F-score is generally better than SL by 0.15. Along with latent visualization, we believe clustered SCL latent space can adopt test data better than SL from this observation.

\begin{table}[ht]
\caption{Audio Classification Accuracy}
\label{tab:audio_acc}
\begin{center}
\begin{tabular}{|c|c|c|c|c|}
\hline
  & Precision & Recall & F-score \\
\hline
% SL (ResNet50) & test & 0.5940 & 0.6749 & 0.6319\\
% \hline
SL (ResNet50)  & 0.5121 & 0.8691 & 0.6444 \\
\hline
% SCL (ResNet50) &  test & \bfseries 0.7101 & \bfseries 0.6774 & \bfseries 0.6934\\
% \hline
SCL (ResNet50)  & \bfseries 0.5913 & \bfseries 0.8687 & \bfseries 0.7036\\
\hline
\end{tabular}
\end{center}
\end{table}

Several factors affect the classifier's performance. Unidirectional microphones can concentrate on only upfront vehicle sound. However, since our model does not separate mixed audio, it can pick up surrounding loud engine sounds and predict them as a false positive even with a directional microphone. These cases cause low precision. Also, we find audio cutout from the wireless microphones affects the classifier's performance. The microphones exhibit intermittent cutout (signal loss) throughout the outdoor recordings. Such signal loss was found to be a common hardware issue for outdoor wireless microphones, even with the most advanced affordable wireless microphone set on the market. Aurally, the cutoff signal has no sound. Digitally, the acquired signal value bounces between specific values, destroying the semantic meaning of the audio samples.

\subsection{IVD Performance}
We perform audio-visual combined evaluation based on our audio validation set. Since our audio validation set is sampled per 1 second (25 frames), we expand combined evaluation set to per entire second (enlarged by 25 times). Each input pair has a 16-frame clip and a 5-second audio sample.
\subsubsection{Qualitative Evaluation}
\begin{figure*}
\centering
\begin{tabular}{cccc}
\includegraphics[width=0.25\linewidth, height=3cm]{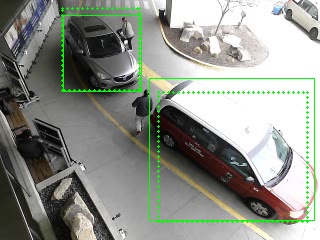} & \includegraphics[width=0.25\linewidth, height=3cm]{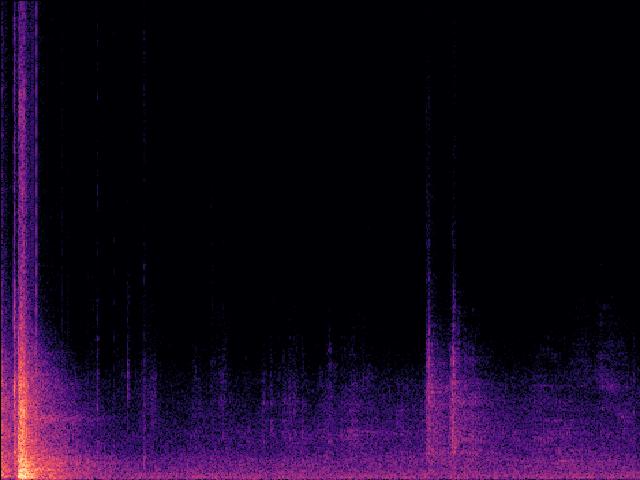} & 
\includegraphics[width=0.25\linewidth, height=3cm]{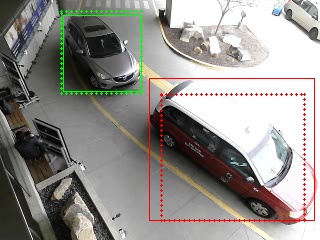} & \includegraphics[width=0.25\linewidth, height=3cm]{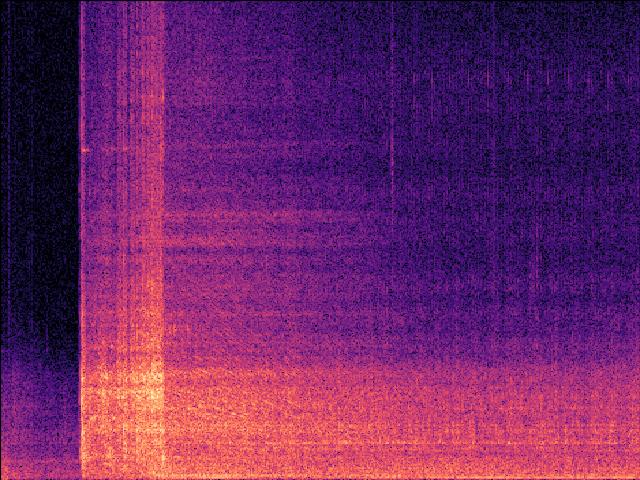} \\ 
\includegraphics[width=0.25\linewidth, height=3cm]{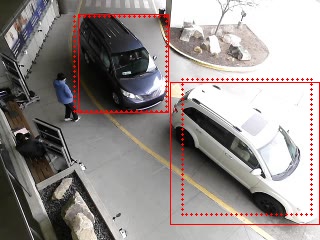} & \includegraphics[width=0.25\linewidth, height=3cm]{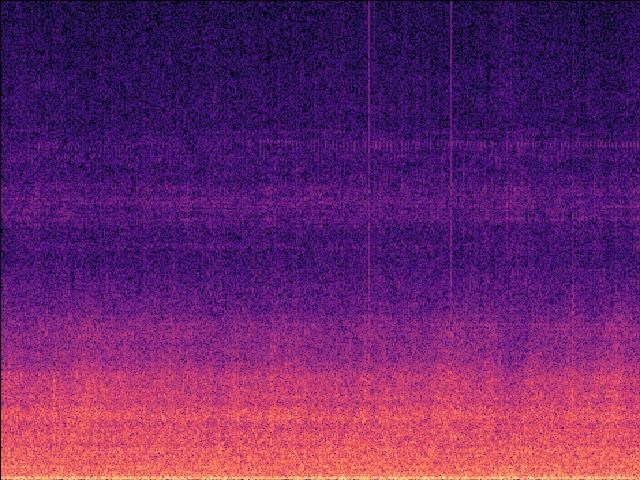} & 
\includegraphics[width=0.25\linewidth, height=3cm]{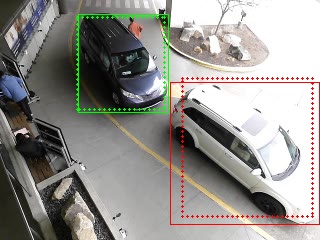} & \includegraphics[width=0.25\linewidth, height=3cm]{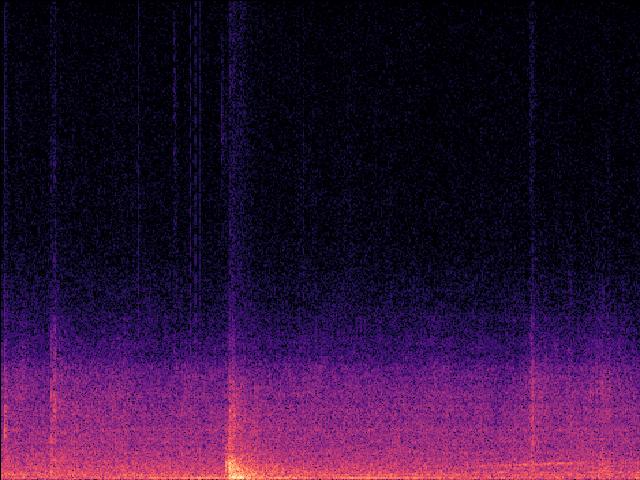} \\ 
\end{tabular}
\caption{IVD Visual Performance. Each row includes detected results and groundtruth annotations along with the corresponding spectrogram. Red, green, and blue bounding boxes are idle, non-idle, and moving labels respectively. Dotted and solid rectangles are prediction and ground truth. }
\label{fig:visual_results}
\end{figure*}

\begin{figure*}
    \centering
    
    \begin{tabular}{c}
        \includegraphics[width=\textwidth]{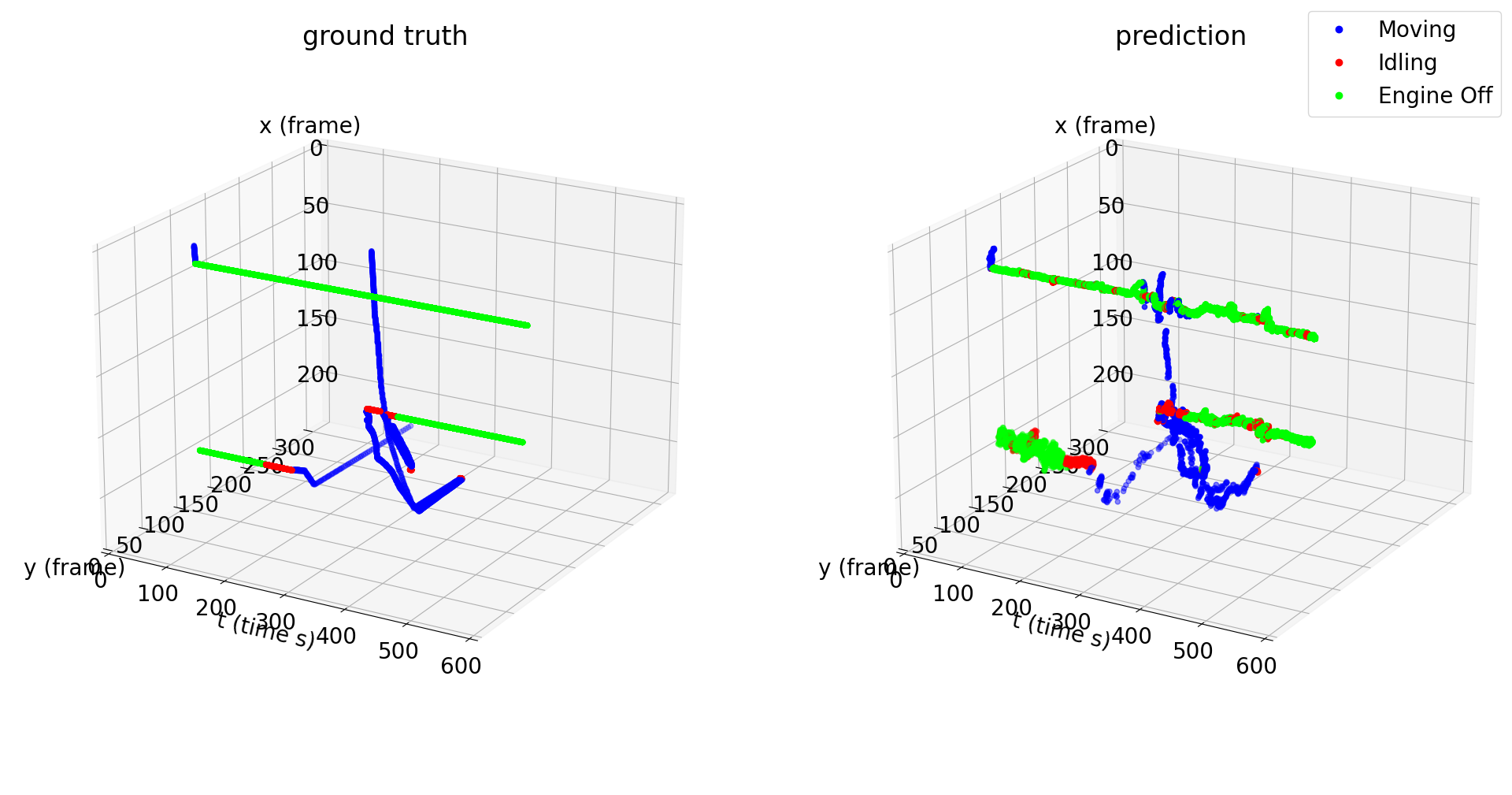} \\
        (a)\\
        \includegraphics[width=\textwidth]{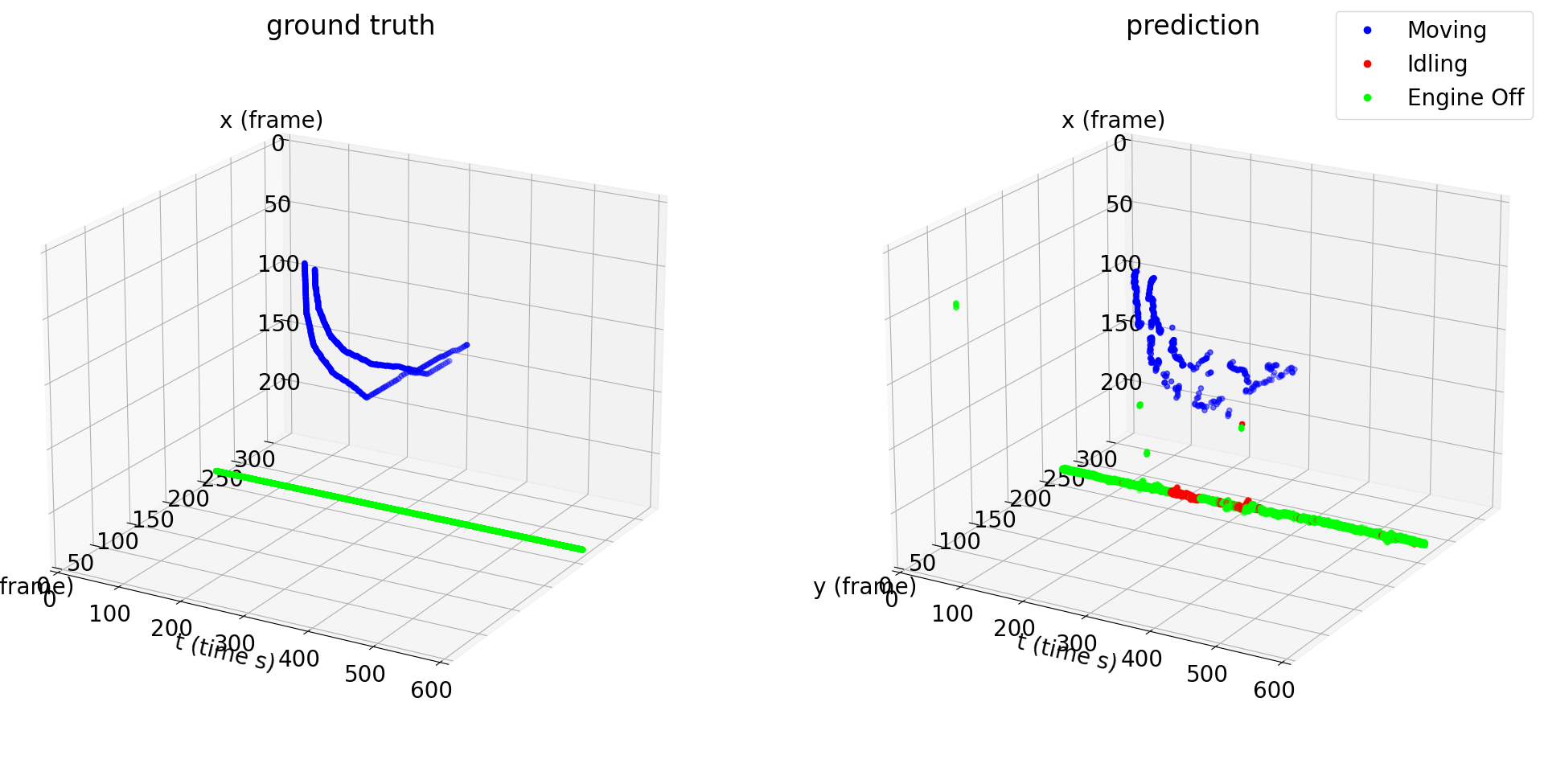} \\
        (b)
    \end{tabular}
    \caption{Two Vehicle Trajectory Reconstruction Examples During Ten-minute Intervals. $x$ and $y$ are centroids of bounding box video frames and $t$ is the time axis. In examples (a) and (b), the shape of the trajectory is the vehicle path through the video sequence. The color of the trajectory indicates the vehicle status, moving, idling, or engine off, of the ground truth and the prediction. In case (a), two vehicles are parked in the target area at the beginning. One of them ignited and drove away. The third vehicle came, parked, and stopped the engine later. Case (b) also includes three vehicles. One was off in the target area, while the other two drove through.}
    \label{fig:trajectory}
\end{figure*}

 Fig. \ref{fig:visual_results} shows the combined audio-visual IVD detection. The trained vehicle motion detector localizes each vehicle very well. As a result, the method is able to find the correct nearest microphone. Additionally, the acquired audio signal is distinct between foreground and background sound. The first row shows two vehicles with the bottom one switching on. The highlighted spectrogram indicates engine ignition has stronger power on frequency bins. The second row shows two vehicles with the upper one switching off, as the corresponding spectrogram turns darker. It also turns out our system is able to handle multiple vehicles with the help of unidirectional microphones.
Two per-frame predictions during 10-minute interval are computed and shown in Fig. \ref{fig:trajectory}. By comparing the color and shape of trajectories, we believe our model is generally capable of capturing correct vehicle position and engine status over a long period of time.

\subsubsection{Quantitative Evaluation}
IVD performance is also evaluated using AP and mAP as three IVD classes. Compared to Table \ref{tab:VDM}, AP Engine Off is similar to AP Stationary. AP Idling is about 10\% drop compared to AP Stationary. We believe this is a reasonable drop by combining audio and visual errors. With similar foreground and background accuracy in the previous section, AP Idling is about 20\% lower than AP Engine Off. Thus, our system struggles more with detecting idling vehicles than with engine-off vehicles, which we believe is due to fewer foreground training and validation samples. Also, our current training set is not big enough to cover a variety of engine sounds, which can lower audio classifier's performance on a different vehicle. However, when threshold IoU is 0.5 and 0.75, our model still has comparable values with models solving other video understanding tasks. 

\begin{table}[ht]
\caption{IVD Performance on Validation Data}
\label{table_example}
\begin{center}
\begin{tabular}{|c|c|c|c|c|}
\hline
 & mAP & AP Moving & AP Idling & AP Engine Off \\
\hline
IoU\textbf{@0.5}  & 80.12 & 78.27 & 71.02 & 91.06 \\
\hline
IoU\textbf{@0.75}  & 30.39 & 25.75 & 17.61 & 47.82 \\
\hline
\end{tabular}
\end{center}
\end{table}

\section{CONCLUSIONS}
In this work, we create a camera and microphone setup and formulate a new problem for IVD. We build an audio-visual algorithm to solve the problem. By deploying the system in real-time for 11 days, it detects IVs in most circumstances and displays smart messages to drivers. We believe this system can be refined and adapted to more real-world scenarios.

\section*{ACKNOWLEDGMENT}
We thank Intermountain Healthcare and LDS Hospital in Salt Lake City, Utah,  United States for providing the field test location for this study.  This material is based upon work supported by the National Science Foundation under Grant No. 1952008  SCC-IRG TRACK 2: SMART AIR: Informing driving behavior through dynamic air-quality sensing and smart messaging.

\bibliographystyle{vancouver}
\bibliography{ios-book-article}

% \begin{thebibliography}{99}

% \bibitem{r1}
% Petitti DB, Crooks VC, Buckwalter JG, Chiu V. Blood pressure levels before dementia.
% Arch Neurol. 2005 Jan;62(1):112-6.

% \bibitem{r2}
% Rice AS, Farquhar-Smith WP, Bridges D, Brooks JW. Canabinoids and pain. In: Dostorovsky JO,
% Carr DB, Koltzenburg M, editors. Proceedings of the 10th World Congress on Pain;  2002 Aug
% 17-22; San Diego, CA. Seattle (WA): IASP Press; c2003. p. 437-68.
% \end{thebibliography}
\end{document}